\documentclass[pmlr]{jmlr}

\usepackage{longtable}

\usepackage{booktabs}
\usepackage[load-configurations=version-1]{siunitx} 

\theorembodyfont{\upshape}
\theoremheaderfont{\scshape}
\theorempostheader{:}
\theoremsep{\newline}

\usepackage{microtype}
\usepackage{graphicx}
\usepackage{booktabs} 
\usepackage{amsmath}
\usepackage{amssymb}
\usepackage{multirow}
\usepackage{caption}
\usepackage{tikz}
\usepackage{natbib}
\usepackage{algorithm}
\usepackage{algorithmic}

\jmlrvolume{86}
\jmlryear{2019}
\jmlrworkshop{Machine Learning for Healthcare}

\title[ASAC: Active Sensing using Actor-Critic models]{ASAC: Active Sensing using Actor-Critic models}

\author{Jinsung Yoon \\
        \addr Electrical and Computer Engineering Department\\
       University of California, Los Angeles, California, United States \\
        jsyoon0823@g.ucla.edu \\
       \AND
        James Jordon \\
        \addr Department of Engineering Science,  University of Oxford, Oxford, United Kingdom\\
        james.jordon@wolfson.ox.ac.uk \\
       \AND
        Mihaela van der Schaar \\
        \addr University of Cambridge, Cambridge, United Kingdom\\
       Alan Turing Institute, London, United Kingdom \\
       University of California, Los Angeles, California, United States \\
        mihaela@ee.ucla.edu}

\begin{document}

\maketitle

\begin{abstract}
Deciding {\em what} and {\em when} to observe is critical when making observations is costly. In a medical setting where observations can be made {\em sequentially}, making these observations (or not) should be an active choice. We refer to this as the active sensing problem. In this paper, we propose a novel deep learning framework, which we call ASAC (Active Sensing using Actor-Critic models) to address this problem. ASAC consists of two networks: a selector network and a predictor network. The selector network uses previously selected observations to determine what should be observed in the future. The predictor network uses the observations selected by the selector network to predict a label, providing feedback to the selector network (well-selected variables should be predictive of the label). The goal of the selector network is then to select variables that balance the cost of observing the selected variables with their predictive power; we wish to preserve the conditional label distribution. During training, we use the actor-critic models to allow the loss of the selector to be ``back-propagated" through the sampling process. The selector network ``acts" by selecting future observations to make. The predictor network acts as a ``critic" by feeding predictive errors for the selected variables back to the selector network. In our experiments, we show that ASAC significantly outperforms state-of-the-arts in two real-world medical datasets.
\end{abstract}

\section{Introduction}\label{sect:intro}
In many medical settings, making observations is costly \cite{weinstein1996cost}. For example, performing lab tests on a patient incurs a cost, both financially as well as causing fatigue to the patient \cite{fatigue, kumwilaisak2008effect}. In such settings, the decision to observe is important. This decision involves a trade-off between the value of the information obtained from the observation and the cost of making the observation. This problem presents itself when the data can be observed sequentially, so that we can observe a particular measurement before deciding which other measurements to observe. This problem presents itself in both static and in time-series settings, with the key difference being that in the time-series setting, the values for a given stream\footnote{We use the term stream to refer to both the sequential values of a time-series variable and the single value of a static variable interchangeably.} will change over time and thus we may wish to re-measure this, whereas in the static setting we know that once we observe a stream, we know its (fixed) value.

Genetic tests, for example, will have the same outcome whether we perform them now or later. As such, it may be advantageous to perform some tests, observe the results, and then decide on further tests to perform based on the results of the first \cite{genetictest}. Because the outcome of the tests will not change over time (we are in a {\em static setting}), there is no need to perform the tests we have already performed again and also no ``worry" that we might miss something by not measuring it now (we can always go back and measure it later).

On the other hand, in an Intensive Care Unit (ICU) setting \cite{ezzie2007laboratory} where important lab tests are being repeated and the results are always changing, we can no longer ignore a stream once it has been measured (its value may have changed since the last time we measured it) and moreover if we decide {\em not} to measure something, then we have missed our chance to measure it in that particular instant (we cannot go back and measure its value in the past). We can, however, still use past observations in determining what to measure next.

We refer to the problem of deciding what to observe in the future based on the measurements observed so far as {\em active sensing} \cite{Active1,Active2}. This problem presents itself in many healthcare applications \cite{Active2,neuro_active}. We formalize the problem of active sensing as a sequential decision making process in which, at each step, we select variables to measure based on all previously selected variables. When selecting variables, we wish to select those which are most predictive of the label, while also minimising cost.

\begin{figure*}
    \centering
    \includegraphics[width =\textwidth]{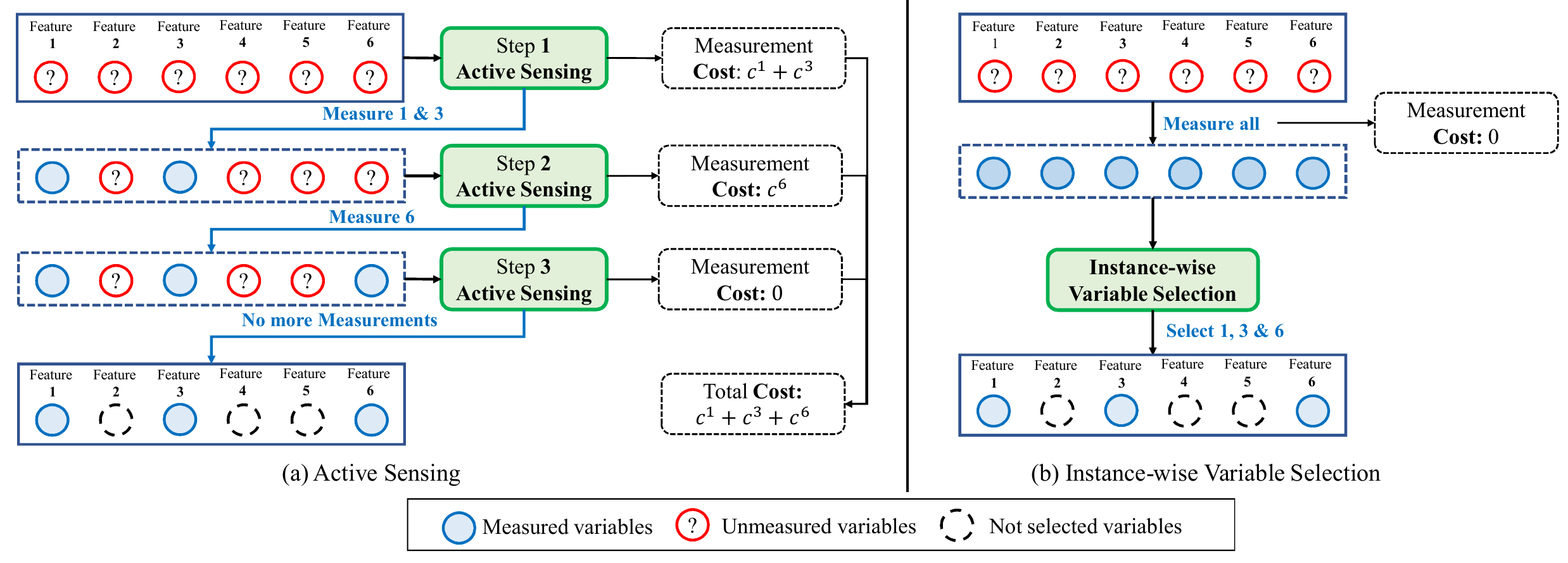}
    \caption{Comparison of Active Sensing and Instance-wise Variable selection in the static setting.}
    \label{fig:active_sensing_framework}
\end{figure*}

This formulation of active sensing is related to instance-wise variable selection frameworks such as \cite{invase,l2x}. In instance-wise variable selection, the goal is to find a minimal subset of variables such that the conditional label distribution is preserved. However, in instance-wise variable selection, all of the variables should be measured {\em before} making the decision of which to select. In such settings, the goal is to efficiently {\em summarize} the information present in the entire feature vector in a lower dimensional feature vector. This is typically because the costly part there is not in observing the value of a variable but rather in presenting the value of the variable. In the active sensing framework, the cost has been shifted from presenting the information to measuring it and as such features that are not selected are not measured. Moreover, in the static setting of instance-wise variable selection, only a single selection is made, whereas for active sensing, both in the static and time-series settings, a sequence of selections is made. Fig. \ref{fig:active_sensing_framework} illustrates these differences between active sensing and instance-wise variable selection. In the Supplementary Materials, we also illustrate these differences in the time-series setting.

\textbf{Technical Significance}
In this paper, we propose ASAC (Active Sensing using Actor-Critic models), an algorithm capable of addressing active sensing in both static and time-series settings. ASAC consists of two networks: a selector network and a predictor network. The selector network uses previously selected features to determine which streams to observe next. The predictor network uses the selected features to predict a label. The networks are trained to minimize a Kullback-Leibler divergence between the conditional label distribution given all features and the conditional label distribution given only the selected features (thus ensuring that the selected features are as predictive of the label as all the features). Cost is introduced by adding a penalty term to the loss. We draw on actor-critic methodology \cite{actor_critic} to allow ``back-propagation" through the sampling process of the selector network. We model each network using LSTMs \cite{LSTM} to deal with sequential inputs and outputs, though any sequential model (e.g. temporal convolutions \cite{wavenet}) could be used.

In our experiments, we demonstrate the efficacy of ASAC in a variety of scenarios using synthetic data. Then, using two real-world medical datasets (ADNI \cite{ADNI} and MIMIC-III \cite{mimic}) we show that ASAC significantly outperforms the existing state-of-the-art methods.

\textbf{Clinical Relevance: } 
In medicine, making observations is usually costly and there is a clear trade-off between cost of measurements (e.g. MRI) and the value of observations (e.g. understanding the patient states based on the observation) \cite{heywang1997contrast}. Therefore, ASAC can do a critical role in clinical decision support that provides advice which observation should be measured and when. The proposed active sensing framework (ASAC) can be widely applied in various medical settings. 

For instance, in an Intensive Care Unit (ICU) setting, there are more than 100 possible measurements on vital signs and lab tests \cite{ezzie2007laboratory} and of course not all the possible tests are necessary for the entire patients. \cite{roberts1993eliminating} Among these extensive combinations of the measurements (and their timings), the proposed methods can provide advice which lab tests and vital signs should be measured and when based on the patient states estimated by the previously measured patients' observations. In breast cancer screening setting, there are various methods to screen the breast cancer such as mammogram, ultrasound, MRI and biopsy \cite{saslow2007american}. Of course, not all the patients need to be screened by all of the above methods \cite{us2009screening}. Most patients only need mammogram to screen the breast cancer and only the subset of the patient needs further screening tools \cite{gotzsche2013screening}. The proposed model can provide the advice in this case as well that which patient needs additional screening examination based on the previous screening results (e.g. mammogram results). 

The proposed methods try to minimize the observation costs without minimum information loss by not observing some measurements. We verify the proposed model in Intensive Care Unit (ICU) setting using MIMIC-III dataset and chronic disease setting using ADNI dataset.

\subsection{Related Works}
This paper draws motivation from existing instance-wise variable selection frameworks such as L2X \cite{l2x}, LIME \cite{lime}, Shapley \cite{shapley}, DeepLIFT \cite{DeepLIFT} and in particular, Instance-wise variable selection (INVASE) \cite{invase}. As noted above, a key difference between instance-wise variable selection and active sensing is in {\em what} is measured {\em before} making the selection. In addition, each of these works formalize the problem only in the static setting (where there are no temporal features). The applications for this problem are restricted to model interpretation and the models cannot be extended to the active sensing framework. 

Deep Sensing \cite{deepsensing} is the work most closely related to ours. Like us, they attempt to solve the active sensing problem using deep learning, especially RNNs. The Deep Sensing framework involves learning 3 different networks: an interpolation network, a prediction network and an error estimation network. Each network is separately optimized for its own objective and then combined together after training to be used for active sensing. On the other hand, ASAC jointly optimizes the selector and predictor networks, {\em both} for the objective of active sensing, doing so by leveraging ideas from actor-critic methods \cite{actor_critic}. Furthermore, Deep Sensing treats each feature independently, deciding what to measure by looking at the affect of a single feature on the label {\em in isolation}. ASAC, on the other hand, jointly estimates the effect of multiple features on the label prediction. This is critical when the features are highly correlated and also when the cost of measuring one feature differs significantly from measuring another noisier correlated feature.
In the experiments, we show that our framework significantly outperforms the state-of-the-art in all settings. Due to space limitations, further details of other related works \cite{attention,transformer,hardattention,Active1,Active2} can be found in the Supplementary Materials.

\section{Problem Formulation}
In this section, we first describe the active sensing problem in the static setting, and then explain the differences in the time-series setting.

\subsection{Static Setting} \label{sec:static}
Let $\mathcal{X} = \mathcal{X}^1 \times ... \times \mathcal{X}^d$ be a $d$-dimensional feature space and $\mathcal{Y}$ be a label space (either $\mathbb{R}$ for regression problems or $\{1,2,...,C\}$ for multi-class classification problems with $C$ classes). We consider random variables $\textbf{X} \in \mathcal{X}$, $Y \in \mathcal{Y}$ with some joint distribution $p$ (and marginal distributions $p_X$ and $p_Y$). For each feature, we assume that there is some cost, $c^i$, where $i=1, ..., d$, associated with measuring the $i$-th feature. The cost vector is denoted as $\textbf{c} = (c^1, ..., c^d)$.

A sensing decision is a vector $\textbf{s} = (s^1, ..., s^d) \in \{0, 1\}^d$ where $s^i = 1$ corresponds to observing $i$-th feature. Let $*$ be any point not in $\mathcal{X}^1, ..., \mathcal{X}^d$. For any sensing vector $\textbf{s}$ and any feature vector $\textbf{x} = (x^1, ..., x^d) \in \mathcal{X}$ let $\textbf{x}(\textbf{s})$ be the vector obtained by
\begin{equation}\label{eq:select}
    x(\textbf{s})^i=\left\{
                \begin{array}{ll}
                  x^i &\text{ if } s^i = 1\\
                  * &\text{ if } s^i = 0\\
                \end{array}
              \right. 
\end{equation}
We refer to $\textbf{x}(\textbf{s})$ as the observed feature vector. In the static setting, we define a sensing decision sequence as $(\textbf{s}_{1}, ..., \textbf{s}_{m})$ where each $\textbf{s}_j$ sensing decision and we require that if $s_{j-1}^i = 1$, then $s_j^i = 1$ so that the sensing decisions form a nested sequence (this is simply so that $\textbf{s}^j$ describes fully which features have already been measured at step $j$) and each $\textbf{s}_{j} = \textbf{s}_{j}(\textbf{x}(\textbf{s}_{j-1}))$ is allowed to depend on $\textbf{x}(\textbf{s}_{j-1})$.

Our goal, then, is to find a sensing decision sequence $(\textbf{s}_{1}, ..., \textbf{s}_{m})$ that minimizes the total cost of measuring the chosen variables (i.e. $\textbf{c}^T\textbf{s}_m = \sum_{i=1}^d c^i \times s_{m}^i$) subject to the conditional distribution of $Y$ given $\textbf{X}$ being equal to the conditional distribution of $Y$ given $\textbf{X}(\textbf{s}_{m})$. That is, we wish to select variables that still allow us to predict $Y$ as well as if we had measured everything, and among the sets of variables that do this, we wish to find the set with minimal measuring cost.

\subsection{Time-series setting}
In the time-series setting, a few modifications need to be made to the problem formulation given in Section \ref{sec:static}. Instead of considering simple random variables, we now consider an indexed family (or sequence) of these random variables $\textbf{X} = (\textbf{X}_{t})_{t \in \mathcal{T}}$ and $(Y_{t})_{t \in \mathcal{T}}$ where $t$ is an index in some time indexing set $\mathcal{T}$ with $\mathcal{T}$ being some bounded subset of either $\mathbb{R}$ or $\mathbb{N}$. In our case we focus on the discrete setting where $\mathcal{T} = \{1, ..., T\} \subset \mathbb{N}$ where $T$ is some {\em random} stopping time (whose distribution we absorb into $p$), with our random processes assumed to be regularly sampled.

In contrast to the static setting, a sensing decision sequence now no longer requires that if $s_{t-1}^i = 1$, then $s_t^i = 1$, since now the values for each component of the process may vary between decisions and so will need to be remeasured if selected again (thus incurring a new cost). In addition, each sensing decision is allowed to depend on {\em all} observations made so far, that is $\textbf{s}_{t} = \textbf{s}_{t}(\textbf{x}(\textbf{s}_{1}), ..., \textbf{x}(\textbf{s}_{t-1}))$.\footnote{This is actually no different to the static setting where $\textbf{x}(\textbf{s}_{j})$ contains all information found in $\textbf{x}(\textbf{s}_{1}), ..., \textbf{x}(\textbf{s}_{j-1})$.} We denote $\textbf{s}_{\leq t} = (\textbf{s}_{1}, ..., \textbf{s}_{t})$, $\textbf{x}_{\leq t} = (\textbf{x}_{1}, ..., \textbf{x}_{t})$ and $\textbf{x}(\textbf{s}_{\leq t}) = (\textbf{x}_1(\textbf{s}_{1}), ..., \textbf{x}_t(\textbf{s}_{t}))$ to simplify notation.

In addition, we can extend this formulation further by allowing measurement delays to be included. Now that we have incorporated a time element, it also becomes natural that some features will take more or less time to measure than others (for example blood cultures can take up to one week to perform). To incorporate this into our formulation, we define a measurement time vector $\tau = (\tau_1, ..., \tau_d) \in \mathcal{T}^d$ which indicates the length of time it takes to measure each feature. Then in this setting, our ``current" feature vector, $\textbf{x}_t(\textbf{s}_{\leq t})$, now depends on all\footnote{In fact, it depends only on $\textbf{s}_u$ for $u \in \{t - \tau_i : i = 1, ..., d\}$, i.e. the times in the past in which measurements were ``started" and whose results would be reported now.} selections made in the past (i.e. on $\textbf{s}_{\leq t}$ rather than just $\textbf{s}_t$) and is defined by
\begin{equation}
x_t(\textbf{s}_{\leq t})^i= \begin{cases}
x_{t - \tau_i}^i &\text{ if } s_{t - \tau_i}^i = 1\\
* &\text{ if } s_{t - \tau_i}^i = 0
\end{cases}
\end{equation}
so that if feature $i$ was selected $\tau_i$ steps ago, then its value appears now in the current set of measured values. In this setting, we also write $\textbf{x}(\textbf{s}_{\leq t}) = (\textbf{x}_1(\textbf{s}_{\leq 1}), ..., \textbf{x}_t(\textbf{s}_{\leq t}))$.

The goal here is as in the static setting, where the total cost is now $\sum_{t=1}^T \textbf{c}^T \textbf{s}_t = \sum_{t=1}^T \sum_{i=1}^d c^i \times s_{t}^i$ and the conditional distribution constraint requires that $Y_{t}$ given $\textbf{X}_{\leq t}$ has the same distribution as $Y_{t}$ given $\textbf{X}(\textbf{s}_{\leq t})$ for all $t \in \{1, ..., T\}$. 

A time-series dataset, which we denote by $\mathcal{D}$, consists of $N$ patient observations, assumed i.i.d. according to $p$ so that $\mathcal{D} = \{(\textbf{x}_{t, i}, y_{t, i})_{t=1}^{T_i}\}_{i=1}^{N}$
where $(\textbf{x}_{t, i}, y_{t, i})_{t=1}^{T_i}$ is the stream corresponding to patient $i$ of (random) length $T_i$.

In the remainder of the paper, the more general time-series setting will be used by default. When reading the rest of the paper, keep in mind that the discussion also applies to the static setting.

\subsection{Optimization problem}\label{sect:optimization}
Based on the above problem formulations, the optimization problem can be determined as follows.
\begin{equation}\label{eq:original_opt}
\begin{aligned}
\min_{\textbf{s}_{1}, ..., \textbf{s}_{T}} \quad & \sum_{t = 1}^{T} \mathbb{E}_{\textbf{x} \sim p_X}  \Big[ \textbf{c}^T \textbf{s}_{t} \Big]\\
\textrm{s.t.} \quad &    (Y_{t}|\textbf{X}_{\leq t} = \textbf{x}_{\leq t}) \,{\buildrel d \over =}\, (Y_{t}|\textbf{X}(\textbf{s}_{\leq t}) = \textbf{x}(\textbf{s}_{\leq t})) \text{ for all } t \in \{1,2,...,T\}\\
\end{aligned}
\end{equation}
In order to find a suitable (tractable) sensing decision sequence, we transform the distributional constraint into a soft constraint using the Kullback-Leibler (KL) divergence. To do this, we consider the problem of minimizing the KL divergence between the two conditional distributions with an added cost penalty term. The objective function we aim to minimize (with respect to the sensing decision sequence) is then
\begin{align}
    \sum_{t=1}^T  \mathbb{E}_{\textbf{x} \sim p_X}\Big[  \big[ &KL(  (Y_{t}|\textbf{X}_{\leq t} = \textbf{x}_{\leq t}) ||
    (Y_{t}|\textbf{X}(\textbf{s}_{\leq t}) = \textbf{x}(\textbf{s}_{\leq t})) ) \big] + \lambda \textbf{c}^T \textbf{s}_{t} \Big]
\end{align}
where $\lambda \geq 0$ is a hyper-parameter that trades-off between the constraint (KL term) and the objective (cost term).

We can rewrite the KL divergence term as
\begin{align*}
    &KL((Y_{t}|\textbf{X}_{\leq t} = \textbf{x}_{\leq t})||(Y_{t}|\textbf{X}(\textbf{s}_{\leq t}) = \textbf{x}(\textbf{s}_{\leq t}))) \\
    =& \int_{\mathcal{Y}} p_Y(y|\textbf{x}_{\leq t})\Big[ \log({p_Y(y|\textbf{x}_{\leq t})}) - \log(p_Y(y|\textbf{x}(\textbf{s}_{\leq t}))) \Big]dy
\end{align*}
and we note that $\log(p_Y(y|\textbf{x}_{\leq t}))$ is independent of the sensing decision sequence $\textbf{s}_{\leq t}$. We can therefore define an equivalent loss, $l(\textbf{x}_{\leq t}, \textbf{s}_{\leq t})$, as follows
\begin{equation} \label{eq:lt}
    l(\textbf{x}_{\leq t}, \textbf{s}_{\leq t})=\int_{\mathcal{Y}} p_Y(y|\textbf{x}_{\leq t})\Big[ - \log(p_Y(y|\textbf{x}(\textbf{s}_{\leq t}))) \Big]dy.
\end{equation}
Then, the new optimization problem is defined as
\begin{equation}\label{eq:new_opt}
\begin{aligned}
\min_{\textbf{s}_1, ..., \textbf{s}_T} \quad & \sum_{t=1}^T\mathbb{E}_{\textbf{x} \sim p_X}\Big[   l(\textbf{x}_{\leq t}, \textbf{s}_{\leq t}) + \lambda \textbf{c}^T \textbf{s}_{t}  \Big].
\end{aligned}
\end{equation}

\section{Proposed model}\label{sect:proposed_model}
In order to solve the optimization problem given in equation (\ref{eq:new_opt}), we first need to estimate the unknown density function: $p_Y(\cdot|\textbf{x}(\textbf{s}_{\leq t}))$. To do this, we introduce a predictor function $f_\phi: \prod_{i=1}^t (\mathcal{X} \times \{0,1\}^d) \rightarrow \mathcal{Y}$ parameterized by $\phi$ which will be trained to predict $y$ given all (selected) observations up until time $t$ (i.e. $\textbf{x}(\textbf{s}_{\leq t})$ and $\textbf{s}_{\leq t}$). 

In order to perform sensing decisions (which are binary), we introduce a selector function $f_\theta: \prod_{i=1}^t (\mathcal{X} \times \{0,1\}^d) \rightarrow [0,1]^d$ parameterized by $\theta$ that will output continuous values in $[0, 1]^d$ which will be treated as probabilities to then be sampled from to create an output in $\{0, 1\}^d$. The selection mechanism is therefore {\em probabilistic} in nature, and as such our optimization problem in (\ref{eq:new_opt}) now needs to include an expectation over the sensing decision sequence $\textbf{s}_{\leq T}$. This selector function $f_\theta$ will take measurements up until time $t$ as input and then output probabilities from which the decision sequence for time $t+1$ will be sampled. In order to ``back-propagate" through the sampling process, we draw on actor-critic models \cite{actor_critic} to derive the gradient of our selector function loss in Section \ref{sec:select}.

These two networks will be trained iteratively. This is important because both functions influence each other. The predictor function directly determines the loss of the selector function and thus has a direct impact on the training of the selector function. The selector function, on the other hand, has the more subtle effect of changing the distribution over which the predictor function needs to perform well. As the selector function is updated, the input distribution for the predictor network changes, and it is important that the predictor function performs well on the new distribution. As such, the predictor network needs to be updated after each selector function update (and vice-versa).

\begin{figure}
    \centering
    \includegraphics[width = 0.8\textwidth]{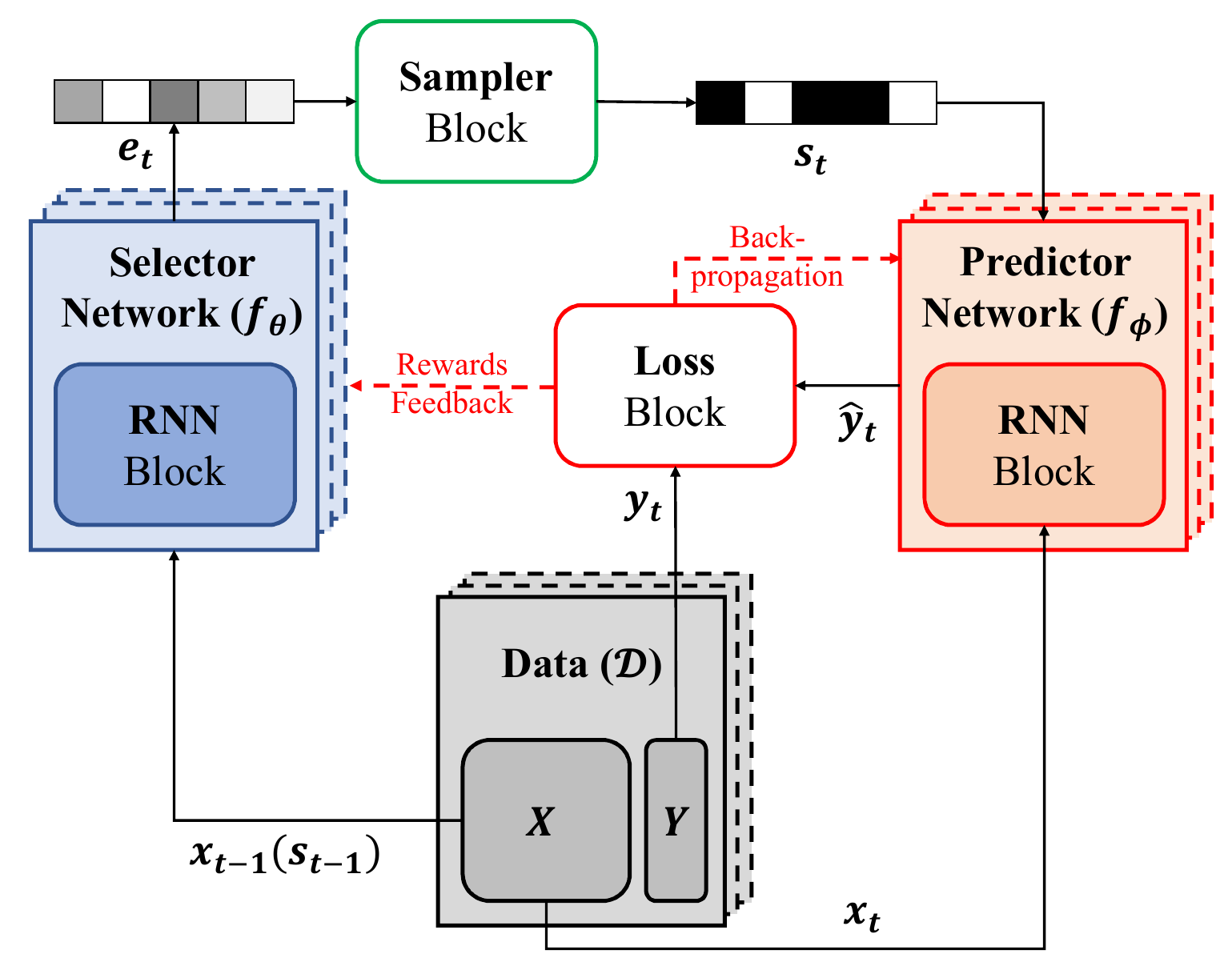}
    \caption{Block diagram of ASAC.}
    \label{fig:block_sensing1}
\end{figure}

\subsection{Predictor function}
The predictor function is trained to minimize a prediction loss
\begin{equation} \label{eq:predloss}
\mathcal{L}(\phi) = \sum_{t=1}^T \mathbb{E}_{\textbf{x} \sim p_X} [l_{t}(\phi)]
\end{equation}
where for $C$-class classification we have the standard cross-entropy loss given by
\begin{align}\label{eq:loss_estimate1}
    \l_t(\phi) = -\sum_{i=1}^C y_t^i \log(f^i_\phi(\textbf{x}(\textbf{s}_{\leq t}), \textbf{s}_{\leq t}))
\end{align}
and for regression we have the standard mean-squared error loss given by
\begin{align}\label{eq:loss_estimate2}
    l_t(\phi) = (y_t - f_\phi(\textbf{x}(\textbf{s}_{\leq t}), \textbf{s}_{\leq t}))^2.
\end{align}
We then use $l_t(\phi)$ as our estimate for $l(\textbf{x}_{\leq t},\textbf{s}_{\leq t})$.

$f_\phi$ can be implemented using any function approximator capable of dealing with time-series inputs (though in the static setting it needs only to be able to deal with static inputs). In this paper, we model $f_\phi$ as a Recurrent Neural Network (RNN) (in particular as an LSTM \cite{LSTM}). 

We explicitly model the predictor function $f_\phi$ using the RNN structure as follows. At time stamp $t$, we first define the hidden state $H_t$ by
$$H_t = f_1(H_{t-1}, \textbf{s}_{t}, \textbf{x}(\textbf{s}_t))$$
where $f_1$ is some function parameterized as a fully connected network (the same network is used for each time point). The output of the predictor network is then given by
$$f_\phi(\textbf{x}(\textbf{s}_{\leq t}), \textbf{s}_{\leq t}) = f_2(H_t) = f_2(f_1(H_{t-1}, \textbf{s}_{t}, \textbf{x}(\textbf{s}_t)))$$
for another function $f_2$ parameterized as a (different) fully connected network.

Note that $H_t$ depends on $H_{t-1}, \textbf{s}_t$ and $\textbf{x}(\textbf{s}_t)$. Iterating this dependency we get that $H_t$ depends on $\textbf{s}_{\leq t}$ and $\textbf{x}(\textbf{s}_{\leq t})$.

\subsection{Selector function} \label{sec:select}
The selector function, $f_\theta: \prod_{i=1}^t (\mathcal{X} \times \{0,1\}^d) \rightarrow [0,1]^d$, outputs probabilities from which we sample independently to obtain a sensing decision. The probability of a given sensing decision, $\textbf{s} = (s^1, ..., s^d)$ given the observations and selections made until time $t$ is given by
\begin{align}\label{eq:pi_explicit}
\pi_\theta(\textbf{s} | \textbf{x}(\textbf{s}_{\leq t}), \textbf{s}_{\leq t}) = \prod_{i=1}^d f_\theta(\textbf{x}(\textbf{s}_{\leq t}), \textbf{s}_{\leq t})^{s^i} (1-f_\theta(\textbf{x}(\textbf{s}_{\leq t}), \textbf{s}_{\leq t}))^{1-s^i} \nonumber
\end{align}
Using a slight abuse of notation, we will write $\textbf{s} \sim \theta$ and $\textbf{s}_t \sim \theta | \textbf{s}_{\leq {t-1}}$ to denote the marginal and conditional distribution of the sensing decision induced by the selector network (note that both of these are conditional on $\textbf{x}_{\leq {t-1}}$). Using this, the objective function in equation (\ref{eq:new_opt}) can be rewritten as follows (we omit the outer expectation ($\mathbb{E}_{\textbf{x} \sim p_X}$) due to space limitation and replace $l(\textbf{x}_{\leq t},\textbf{s}_{\leq t})$ with $l_t(\phi)$):
\begin{align}
\mathcal{L}(\theta) &= \sum_{t = 1}^{T}\mathbb{E}_{\textbf{s} \sim \theta}\big[l_t(\phi) + \lambda \textbf{c}^T\textbf{s}_t\big] \\
&= \sum_{t = 1}^{T}\mathbb{E}_{\textbf{s}_1 \sim \theta}  \Big[ \cdots \mathbb{E}_{\textbf{s}_t \sim \theta|\textbf{s}_{\leq t-1}} \big[l_t(\phi) + \lambda \textbf{c}^T\textbf{s}_t\big] \Big] \nonumber \\
&=  \sum_{t = 1}^T \sum_{\textbf{s}_1 \in \{0,1\}^d}  \pi_\theta(\textbf{s}_1) \Big[  \sum_{\textbf{s}_t \in \{0,1\}^d} \pi_\theta(\textbf{s}_t|\textbf{s}_{\leq t-1}) 
 \times \big[l_t(\phi) + \lambda \textbf{c}^T\textbf{s}_t\big] \Big] \nonumber \\
&= \sum_{t = 1}^T \sum_{\textbf{s}_{\leq t} \in \{0,1\}^{d \times t}}    [\prod_{\tau=1}^t \pi_\theta(\textbf{s}_\tau|\textbf{s}_{\leq \tau-1}) ] \big[l_t(\phi) + \lambda \textbf{c}^T\textbf{s}_t\big]\nonumber
\end{align}
Using ideas from actor-critic models \cite{actor_critic} (details can be found in the Supplementary Materials), the gradient of this loss $\nabla_{\theta}\mathcal{L}(\theta)$ can be shown to be
\begin{equation} \label{eq:loss}
\nabla_{\theta}\mathcal{L}(\theta) = \sum_{t = 1}^T \sum_{j=1}^t \mathbb{E}_{\textbf{s} \sim \theta}  \big[[l_t(\phi) + \lambda \textbf{c}^T\textbf{s}_t] \nabla_\theta \log \pi_\theta(\textbf{s}_j|\textbf{s}_{\leq j-1}) \big]
\end{equation}
where $\nabla_{\theta} \log \pi_\theta(\textbf{s}_j|\textbf{s}_{\leq j-1})$ is
\begin{align} \label{eq:grad}
\sum_{i=1}^d \Big[ s_j^i \nabla_{\theta} \log f_\theta^i(\textbf{x}(\textbf{s}_{\leq j-1}), \textbf{s}_{\leq j-1}) - (1-s^i_j) \nabla_{\theta} \log f_\theta^i(\textbf{x}(\textbf{s}_{\leq j-1}), \textbf{s}_{\leq j-1}) \Big].
\end{align}
which can be directly deduced from Equation (\ref{eq:pi_explicit}).

We explicitly model the selector function $f_\theta$ using the RNN structure as follows. At time stamp $t$, we first define the hidden state $h_t$ by
$$h_t = f_3(h_{t-1}, \textbf{s}_{t}, \textbf{x}(\textbf{s}_t))$$
where $f_3$ is some function parameterized as a fully connected network (the same network is used for each time point). The output of the selector network is then given by
$$f_\theta(\textbf{x}(\textbf{s}_{\leq t}), \textbf{s}_{\leq t}) = f_4(h_t) = f_4(f_3(h_{t-1}, \textbf{s}_{t}, \textbf{x}(\textbf{s}_t)))$$
for another function $f_4$ parameterized as a (different) fully connected network.

Note that $h_t$ depends on $h_{t-1}, \textbf{s}_t$ and $\textbf{x}(\textbf{s}_t)$. Iterating this dependency we get that $h_t$ depends on $\textbf{s}_{\leq t}$ and $\textbf{x}(\textbf{s}_{\leq t})$.

\begin{figure}[t!]
    \centering
    \includegraphics[width = 0.6\textwidth]{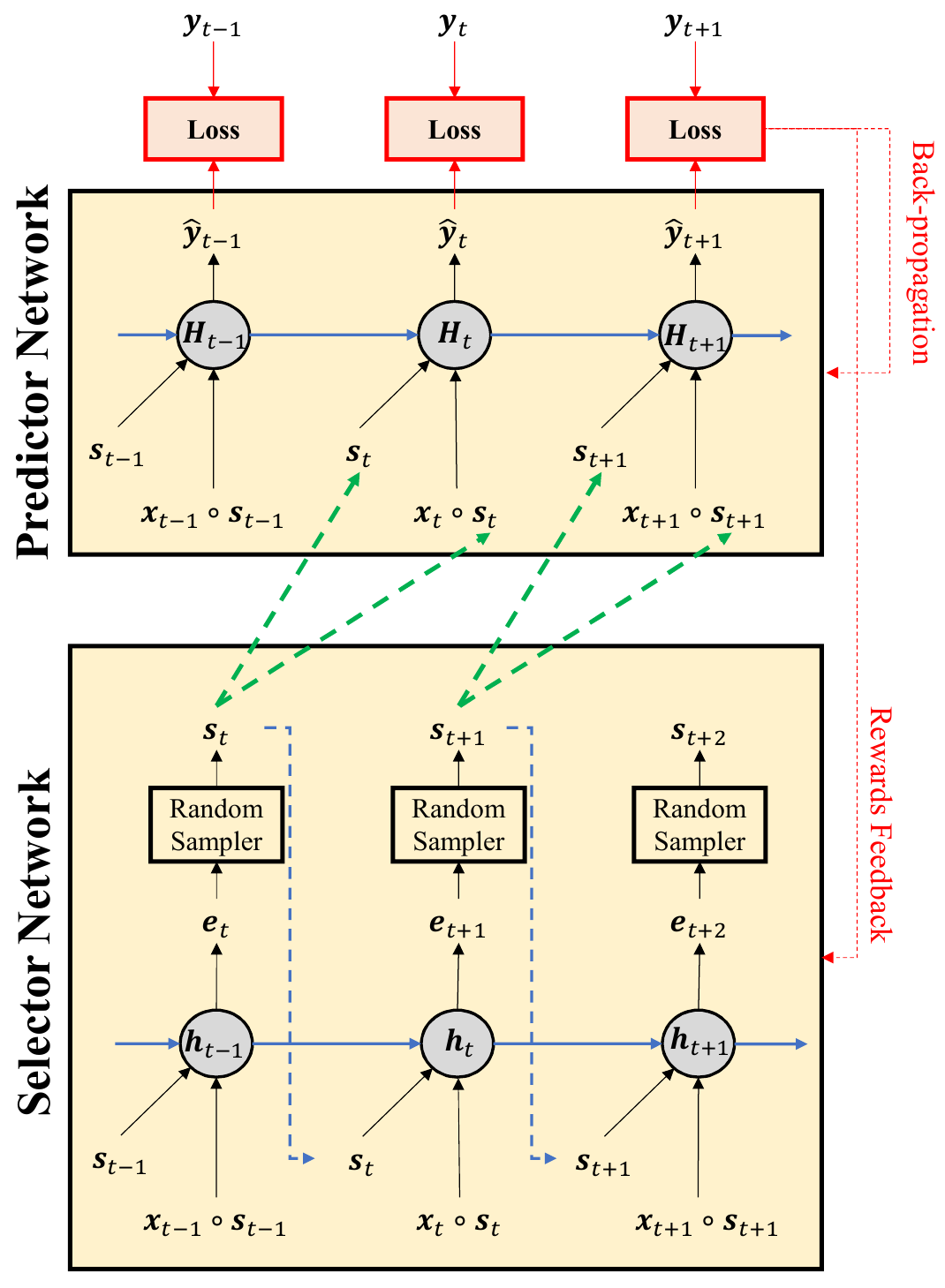}
    \caption{Block diagram of ASAC in a time-series setting.}
    \label{fig:block_sensing2}
\end{figure}

Fig. \ref{fig:block_sensing1} illustrates the entire structure of ASAC. Fig. \ref{fig:block_sensing2} illustrates ASAC in the time-series setting. The lower part of Fig. \ref{fig:block_sensing2} depicts the selector network ($e_t$ represents the output of $f_\theta$ at time stamp $t$) and the upper part of the Fig. \ref{fig:block_sensing2} depicts the predictor network.

\subsection{Training the networks} \label{sec:training}
The selector and predictor networks are jointly and iteratively trained. First, the predictor network ($f_\phi$) is trained to minimize the predictor loss $\mathcal{L}(\phi)$ given the sensing decisions made by the selector network ($f_\theta$). We investigated the effect of sampling multiple sensing decisions for the same time-step and sample but found that this had very little effect on the performance. As such, when we create a mini-batch to train the predictor network with, we sample only 1 sensing decision for each sample in the mini-batch.

The parameters of the predictor network are updated according to
\begin{align*}
	\phi &\gets \phi - \beta \frac{1}{{n_{mb}}} \sum_{i=1}^{n_{mb}} \sum_{t=1}^{T_i} (y_{t,i} - f_\phi(\textbf{x}(\textbf{s}_{\leq t, i}), \textbf{s}_{\leq t, i}))^2
\end{align*}
where $n_{mb}$ is the size of the mini-batch and $\beta>0$ is the learning rate (specific to the predictor network). Then, given a fixed predictor network, the selector network parameters are updated according to
\begin{align*}
    \theta \gets \theta - \alpha \frac{1}{{n_{mb}}}  &\sum_{i=1}^{n_{mb}}  \sum_{t = 1}^{T_i} \sum_{\tau=1}^t  \Big[ \big[l_{t, i}(\phi) + \lambda \textbf{c}^T\textbf{s}_{t,i}\big] \times\nabla_\theta \log  f_\theta(\textbf{x}(\textbf{s}_{\leq \tau, i}),\textbf{s}_{\leq \tau, i}) \Big]
\end{align*}
where $\alpha>0$ the learning rate (specific to the selector network). Pseudo-code can be found in the Supplementary Materials.

\subsubsection{Missing Data during Training} \label{sec:missing}
The loss we have derived lends itself naturally to missing data in the training set. By inspecting Equations (\ref{eq:loss}) and (\ref{eq:grad}), we see that the gradient is made up of a sum over each feature. During training, when ``back-propagating" to the selector network, for features that were selected by the network but were missing (and so their measurement can't be given), we do not back-propagate their loss. The selector network only back-propagates for both not-selected features and selected-and-not-missing features.

\section{Experiments}\label{sect:experiments}

\subsection{Data Description}
We use two real-world medical datasets to evaluate the performance of ASAC against Deep Sensing \cite{deepsensing} for various cost constraints. 

\textbf{ADNI dataset:} The Alzheimer’s Disease Neuro-imaging Initiative (ADNI) study data is a longitudinal survival dataset of per-visit measurements for 1,737 patients \cite{ADNI}. The data tracks disease progression through clinical measurements at 1/2-year intervals, including quantitative biomarkers, cognitive tests, demographics, and risk factors. For this dataset, the adverse event we predict is unstable state occurrence.

\textbf{MIMIC-III dataset:} The MIMIC-III dataset \cite{mimic} has de-identified electronic health records (EHR) from Beth Israel Deaconess Medical Center from 2001 to 2012. It was collected from two information systems (Philips CareVue Clinical and iMDsoft MetaVision ICU) that have very different data structures. We only use data collected by MetaVision (after 2008) for consistency. We extract 40 physiological data streams from lab tests (20) and vital signs (20) that have the lowest missing rates (including heart rate, respiratory rate, blood pressures). The number of patients is 23,153 and there are 5,143 sequences of length larger than 100 time steps with the longest being 1,487 time steps.
For this dataset, the adverse event we predict is death.

\begin{figure*}[t]
	\centering
	\includegraphics[width=\textwidth]{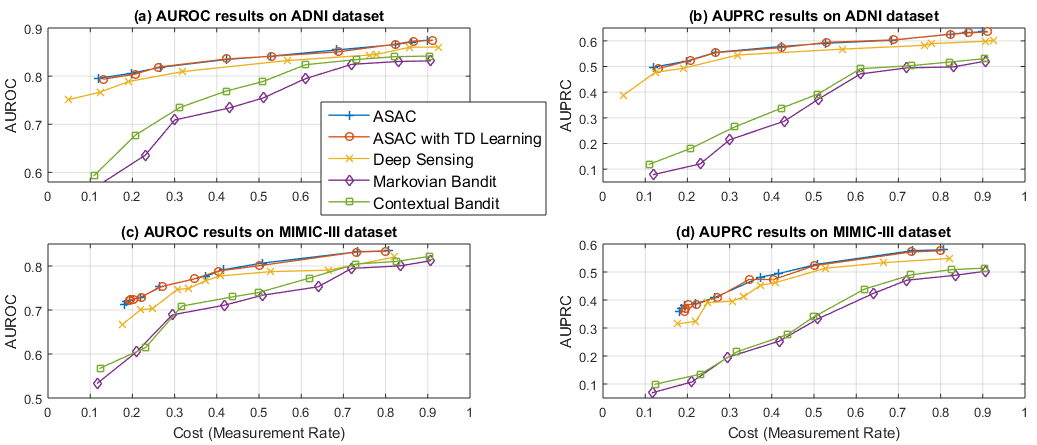}
	\caption{Results on risk predictions on both ADNI and MIMIC-III dataset with various cost constraints in terms of AUROC and AUPRC. X-axis is cost constraints (rate of selected measurements). Y-axis is predictive performance.}
	\label{fig:realresult}
\end{figure*}

\subsection{Experimental Results}
We evaluate the performance of ASAC against 3 benchmarks: (1) Deep Sensing \cite{deepsensing}, (2) Contextual Bandit \cite{contextualbandit1,contextualbandit2}, (3) Markovian Bandit \cite{markovbandit}. Furthermore, we also evaluate our model when replacing the actor-critic methodology with TD learning \cite{tdlearning} and refer to this model as ASAC with TD learning. We randomly divided the dataset into mutually exclusive training (80\%) and testing (20\%) sets. We conducted 10 independent experiments with different training/testing sets in each and we report the mean and standard deviation of the performance in the 10 experiments.

In Fig. \ref{fig:realresult}, we plot AUROC and AUPRC against the average measurement rate of all features (corresponding to all features being assigned the same cost). In MIMIC-III, we ignore the cost when a missing feature is selected.

As can be seen in Fig. \ref{fig:realresult}, ASAC (and ASAC with TD learning) consistently outperforms all 3 benchmarks, achieving higher predictive power for the same cost across all costs. Variance analysis can be found in the Supplementary Materials showing that ASAC achieves statistically significant improvements over all 3 benchmarks. We see from Fig. \ref{fig:realresult}(c)(d) that ASAC is robust to missing data, where we note that around 40\% of the data is missing in the MIMIC-III dataset. ASAC and ASAC with TD learning achieve similar performances indicating that the ASAC framework can be robustly combined with various Reinforcement Learning frameworks to address the active sensing problem.

We can see a trade-off between accuracy and observational costs. In the ASAC framework, we can either maximize the accuracy given constraints on observational costs or minimize the cost given the desired accuracy constraint. As can be seen in the grid line in Figure \ref{fig:realresult}; the horizontal line represents fixing the accuracy, vertical line represents fixing the cost. We illustrate these trade-off curves for ASAC in Figure \ref{fig:realresult}, which shows that ASAC outperforms state-of-the-art under both types of constraints.

\subsection{Analysis on ASAC with Synthetic datasets}
We perform 3 synthetic experiments that we believe capture key attributes of an active sensing method. In each simulation, the feature distribution is a 10 dimensional auto-regressive Gaussian model over 10 time steps, i.e.
\begin{equation}
\textbf{X}_t = {\phi} \odot \textbf{X}_{t-1} + (1-{\phi}) \odot \textbf{Z}_t
\end{equation}
where $\odot$ denotes element-wise multiplication, $\phi \in [0, 1]^{10}$ is a vector that determines the dependency of each feature on the past (a higher $\phi$ corresponds to a larger dependency on the past) and $\textbf{Z}_t$ is an independent Gaussian noise vector $\textbf{Z}_t \sim \mathcal{N}(0, \textbf{I}^{10})$.

\subsubsection{Time dependency vs Measurement rate}
In our first experiment, we investigate the effect of time dependency on measurement rate of a variable. If we fix the cost and label-dependency of all variables to be the same, then we would expect a variable with a large $\phi$ to be measured less frequently by a good active sensing method (due to being more easily predicted from previous values).

To do this, we set the label, $Y_t$ according to
\begin{equation}
    Y_t = \exp(-0.1 \times |\sum_{i=1}^{10} X^i_t|) + \epsilon
\end{equation}
where $\epsilon \sim \mathcal{N}(0,0.1)$. We set the cost for each variable to be the same, which we vary from $1$ to $5$. We set $\phi = (0,0.1,...,0.9)$. The measurement rate (the selection probability) of each variable is reported in Table \ref{tab:synthetic_1}, along with the overall RMSE for each experiment.

\begin{table}[h!]
    \centering
    \begin{tabular}{c|c|c|c|c|c}
    \toprule
        $\phi^i$/Cost  & 1 & 2 & 3 & 4 & 5 \\
        \midrule
        0 ($\textbf{X}^1_t$)  & 1.00& 1.00& 1.00& 0.46& 0.38\\
        0.1 ($\textbf{X}^2_t$)  &1.00& 1.00& 1.00& 0.44&  0.36\\
        0.2 ($\textbf{X}^3_t$)  & 1.00& 1.00& 1.00& 0.30& 0.26 \\
        0.3 ($\textbf{X}^4_t$)  &  1.00& 1.00& 1.00& 0.25& 0.12 \\
        0.4 ($\textbf{X}^5_t$)  & 1.00& 1.00& 1.00& 0.22& 0.10 \\
        0.5 ($\textbf{X}^6_t$)  &  1.00& 0.98& 0.23& 0.21& 0.07 \\
        0.6 ($\textbf{X}^7_t$)  & 1.00& 0.94& 0.13& 0.10& 0.05 \\
        0.7 ($\textbf{X}^8_t$)  & 1.00& 0.93& 0.07& 0.03& 0.01 \\
        0.8 ($\textbf{X}^9_t$)  & 0.92& 0.41& 0.03& 0.02& 0.0 \\
        0.9 ($\textbf{X}^{10}_t$)  & 0.45& 0.11& 0.01& 0.01& 0.0 \\
        \midrule
        RMSE & 0.106& 0.110& 0.126& 0.138& 0.146 \\
        \bottomrule
    \end{tabular}
    \caption{Measurement rate of each feature when each feature has a different auto-regressive coefficient.}
    \label{tab:synthetic_1}
\end{table}

As can be seen in Table \ref{tab:synthetic_1}, ASAC meets our expectations. Features with a low $\phi$, are regularly re-measured since past values are not as predictive of the present value, whereas features with a high $\phi$ are measured less frequently. As cost increases, we also see a monotonic decrease in the measurement rate of all variables.

\subsubsection{Cheaper but noisier features} \label{sec:synth2}
In our second synthetic experiment, we investigate the effect of having cheaper, noisier versions of our original 10 features. In this experiment we are interested in understanding how well ASAC can trade-off between the cost and noise level of the noisy versions. This setting has several real-world parallels; in medicine, cheap at-home tests (such as blood pressure tests and home pregnancy tests) exist, but are less reliable (noisier) than the more expensive state-of-the-art procedures that would be used in, say, a hospital setting.

To model this, we introduce 10 new noisy features
\begin{equation}
    \hat{\textbf{X}}_t = \textbf{X}_t + \delta
\end{equation}
where $\delta \sim \mathcal{N}(0,\gamma)$ with $\gamma>0$ controlling the ``noisiness". In this experiment, we set the label according to
\begin{equation}
Y_t = \exp(-|0.1X^1_t+0.2X^2_t+0.3X^3_t+0.4X^4_t|) + \epsilon
\end{equation}
where now we have set different magnitudes for the coefficients of the first 4 variables (and the last 6 variables are now just there as pure noise). We would expect that as we increase the cost of the true variables (or equivalently decrease the cost of the noisy variables), the variables with lower importance ($X^1$ and $X^2$) will be the first ones to be ``replaced" with their noisy version, whereas it will take a higher cost for $X^4$ to be replaced with $\hat{X}^4$. 

We fix the cost of the original features to be 1, and investigate noise levels $\gamma \in \{0.2, 0.4, 0.6\}$ and vary the cost of a noisy feature to be $\hat{c} \in \{0.1.0.2.0.5\}$. We set $\phi^i = 0.5$ for all $i$. In Table \ref{tab:synthetic_2} we report the measurement rate of each of the first 4 variables and their noisy versions. 

\begin{table}[h!]
    \centering
    \begin{tabular}{c|c|cc|cc|cc}
    \toprule
       \multirow{2}{*}{$\gamma$} & Cost & 1 & 0.1 &   1 & 0.2&   1 & 0.5 \\
       \cmidrule{2-8}
       & Features & $\textbf{X}_t$ & $\hat{\textbf{X}}_t$& $\textbf{X}_t$ & $\hat{\textbf{X}}_t$& $\textbf{X}_t$ & $\hat{\textbf{X}}_t$\\
        \midrule
       \multirow{5}{*}{0.2} & ${X}^1$ & 0.00&1.00 & 0.00&1.00 &0.00 &1.00  \\
       & ${X}^2$ & 0.00&1.00 &0.00 &1.00 &0.29 &0.70  \\
       & ${X}^3$ & 0.00&1.00 &0.00 &1.00 &1.00 &0.00  \\
       & ${X}^4$ & 0.00&1.00 &0.00 &1.00 &1.00 &0.00  \\
        \midrule
       \multirow{5}{*}{0.4} & ${X}^1$ &0.00 & 1.00& 0.00&0.94 &1.00 & 0.00 \\
       & ${X}^2$ & 0.00& 1.00& 0.30& 0.65 & 1.00& 0.00 \\
       & ${X}^3$ & 1.00& 0.00& 1.00& 0.00&1.00 & 0.00 \\
       & ${X}^4$ & 1.00& 0.00& 1.00& 0.00&1.00 & 0.00 \\
        \midrule
       \multirow{5}{*}{0.6} & ${X}^1$ & 0.00 &1.00 & 0.51& 0.33& 1.00& 0.00 \\
       & ${X}^2$ & 0.75 & 0.25& 1.00& 0.00& 1.00& 0.00  \\
       & ${X}^3$ & 1.00& 0.00& 1.00& 0.00& 1.00& 0.00 \\
       & ${X}^4$ & 1.00& 0.00& 1.00& 0.00& 1.00& 0.00 \\
        \bottomrule
    \end{tabular}
    \caption{Measurement rate based on different cost and noise parameter $\gamma$ for original feature ($\textbf{X}_t$) and noisy feature ($\hat{\textbf{X}}_t$).}
    \label{tab:synthetic_2}
\end{table}

As can be seen in Table \ref{tab:synthetic_2}, ASAC meets our expectations. As we move right and down in the table (corresponding to an increasing cost for the noisy feature and increasing noise, respectively), we see that true features are selected more frequently but that the noisy versions for the less predictive features ($X^1$ and $X^2$) are sometimes selected even at higher costs and noise levels. In particular, at $(\gamma,\hat{c}) = (0.2, 0.2)$, only the noisy features are selected. When $\gamma$ is increased to 0.4, ASAC starts to select $X^3$ and $X^4$ all of the time, and $X^2$ some of the time, while the noisy version of $X^1$ is always preferred. When we increase $\gamma$ to 0.6, the true version of $X^2$ is the only version selected by ASAC and the true version of $X^1$ finally becomes desirable enough to measure (sometimes).

\subsubsection{$Y$ dependent cost}
\begin{table}[h!]
	\centering
	\begin{tabular}{c|c|c|c|c|c|c}
		\toprule
		$\eta  $  & \multicolumn{2}{c|}{0.1} & \multicolumn{2}{c|}{0.3} & \multicolumn{2}{c}{0.5}  \\
		\midrule
		Features  & $\textbf{X}_t$ & $\hat{\textbf{X}}_t$ & $\textbf{X}_t$ & $\hat{\textbf{X}}_t$ & $\textbf{X}_t$ & $\hat{\textbf{X}}_t$\\
		\midrule
		 ${Y}_t=1$ & 0.89 & 0.10& 0.63 & 0.21& 0.25 & 0.69  \\
		 ${Y}_t=0$ & 0.13 & 0.81& 0.14 &  0.80& 0.12 & 0.78\\
		\bottomrule
	\end{tabular}
	\caption{Measurement rate when the cost is different for $Y_t = 1$ and $Y_t = 0$.}
	\label{tab:synthetic_3}
\end{table}

In our final synthetic experiment, we allow for a cost that depends on $Y$. In our medical example, this could correspond to the fact that when a patient is sick, it is more important to be sure about it, than when a patient is well. In the presence of the cheaper-but-noisier features from \ref{sec:synth2}, we expect a worsening condition to create a switch in selections. While a patient is healthy, we are happy to monitor the patient using the at-home tests, but when a patients condition appears to be deteriorating, it becomes more important that accurate measurements are made than cost being kept low.

We model this by incorporating the patients condition into the cost, setting the cost when the patient is sick ($Y_t = 1$) to be $\eta \in [0, 1]$ times the cost when the patient is healthy ($Y_t = 0$)\footnote{By reducing the measurement cost, we are equivalently up-weighting the importance of accurately predicting.}. We investigate $\eta \in \{0.1, 0.3, 0.5\}$. 

We generate the true features as before, now with $\phi^i = 0.9$ for all $i$, and generate noisy features as in \ref{sec:synth2} with $\gamma = 0.4$. We set the label to be binary according to
\begin{equation*}
        Y_t = \begin{cases}
                  1 \text{, w.p } \exp(-0.1 \times |\sum_{i=1}^{10} X^i_t + \epsilon - 2|) \\
                  0 \text{, w.p } 1-\exp(-0.1 \times |\sum_{i=1}^{10} X^i_t + \epsilon - 2|)
                \end{cases}
\end{equation*}
where ``w.p" means ``with probability".

We see from Table \ref{tab:synthetic_3} that ASAC is able to correctly identify that measuring the true features is more important when $Y_t = 1$, with measurement frequencies while $Y_t = 1$ for the true features being higher for all 3 values of $\eta$. When $\eta = 0.5$, which corresponds to the cost being half as important while the patient is sick, we see that true features are measured twice as frequently. As $\eta$ decreases, and so accurate predictions become more important, we see that ASAC selects true features more frequently. When $\eta=0.1$, ASAC selects true feature nearly 7 times more frequently while the patient is sick compared to when they are not. ASAC can therefore be used to handle settings where the trade-off between measurement cost and prediction accuracy varies according to the label (which is often the case in medicine).

\section{Conclusion}
We propose a novel active sensing framework, called Active Sensing using Actor-Critic models (ASAC), to address the important question of {\em what} and {\em when} to observe. This is critical when observations are costly. We demonstrated through real-world and synthetic experiments that the ASAC framework can significantly reduce the cost of observation with only a small loss in predictive power. Using the MIMIC-III dataset we also demonstrated that ASAC is robust to missing data.

We believe ASAC has wide-ranging applications, both in cost reduction but also for things such as planning, in which patients can be told when they might expect to need their next check-up and for what (i.e. personalized screening).


\end{document}